\documentclass{article}

\usepackage{arxiv}

\usepackage[utf8]{inputenc} 
\usepackage[T1]{fontenc}    
\usepackage{hyperref}       
\usepackage{url}            
\usepackage{booktabs}       
\usepackage{amsfonts}       
\usepackage{nicefrac}       
\usepackage{microtype}      
\usepackage{lipsum}

\usepackage{graphicx}
\usepackage{subfigure}

\title{Learning New Tricks from Old Dogs -- Inter-Species, Inter-Tissue Domain Adaptation for Mitotic Figure Assessment}

\author{
  Marc Aubreville \\
  Pattern Recognition Lab\\
  Computer Sciences\\
  Friedrich-Alexander-Universität Erlangen-Nürnberg\\
  Erlangen, Germany\\
   \And
 Christof A. Bertram \\
  Institute of Veterinary Pathology\\
  Freie Universit\"at Berlin\\
  Berlin, Germany \\
  \And
  Samir Jabari \\
  Institute of Neuropathology\\
  Friedrich-Alexander-Universit\"at Erlangen-N\"urnberg\\
  Erlangen, Germany
  \AND
  Christian Marzahl\\
  Pattern Recognition Lab\\
  Computer Sciences\\
  Friedrich-Alexander-Universität Erlangen-Nürnberg\\
  Erlangen, Germany\\
   \And
  Robert Klopfleisch \\
  Institute of Veterinary Pathology\\
  Freie Universit\"at Berlin\\
  Berlin, Germany \\
  \And
  Andreas Maier\\
  Pattern Recognition Lab\\
  Computer Sciences\\
  Friedrich-Alexander-Universität Erlangen-Nürnberg\\
  Erlangen, Germany\\
}

\begin{document}
\maketitle
%

\newcommand{\etal}{et al.~}

\begin{abstract}
For histopathological tumor assessment, the count of mitotic figures per area is an important part of prognostication. Algorithmic approaches - such as for mitotic figure identification - have significantly improved in recent times, potentially allowing for computer-augmented or fully automatic screening systems in the future. This trend is further supported by whole slide scanning microscopes becoming available in many pathology labs and could soon become a standard imaging tool.

For an application in broader fields of such algorithms, the availability of mitotic figure data sets of sufficient size for the respective tissue type and species is an important precondition, that is, however, rarely met. While algorithmic performance climbed steadily for e.g. human mammary carcinoma, thanks to several challenges held in the field, for most tumor types, data sets are not available. 

In this work, we assess domain transfer of mitotic figure recognition using domain adversarial training on four data sets, two from dogs and two from humans. We were able to show that domain adversarial training considerably improves accuracy when applying mitotic figure classification learned from the canine on the human data sets (up to +12.8\% in accuracy) and is thus a helpful method to transfer knowledge from existing data sets to new tissue types and species.
 
\end{abstract} 

\section{Introduction}
Mitotic figures, i.e. cells undergoing cell devision, are an important marker for tumor proliferation and their density is used in many tumor grading schemes for prognostication \cite{2759-01}.
Fostered by the availability of datasets, this field has seen significant algorithmic advances in the very recent past. Especially in the field of human mammary carcinoma, one of the most common tumors in women, data sets like MITOS~\cite{2759-03} were able to include a great deal of variance that are typical in pathology. 
Since mitosis is a multi-phase process, the visual variability of its appearance in microscopy images is high. Besides this, there is a number of other factors that increase variance: staining is a manual or semi-automated process, and such is the preparation of the tissue sections to put on the microscopy slide. 
Another source of variance is the tissue itself, as especially macroscopic structures are differing considerably between tissue types. On top of that, we can expect differences between humans and other species - however, it could be well debated if, given a significant sources of variability, these would always play a major role.

All these factors cause a domain shift between data sets that is well-known in digital histopathology, and thus machine learning algorithms that have been trained on one data set will rarely directly transfer to another. Two classes of algorithmic adaptations to models are typically used in this regard: First, we can train a model on a large data set in one domain, which will allow to provide the model with a good initialization of its feature space for the next task, which is training the model with a (smaller) set of images in the target domain - typically referred to as \textit{transfer learning}. This, however, requires annotated images in the target domain, which are also not always available to a sufficient amount.

Another approach is domain adaptation, which can also be done unsupervised, i.e. without any annotation and just images available in the target domain. These approaches rely solely on statistical properties of the input distributions, and will only work sufficiently if the derived real labels of the respective known source domain and the (assumed to be unknown) target domain follow the same ruleset. 
 This is especially tricky in the field of mitotic figure detections, where individual thresholds play a significant role and thus the labels are significantly dependent on the expert defining the label.

Recently, Li \textit{et al.} have shown, that a combination of an object detection network and a refining second stage classifier \cite{2759-04} show superior performance in mitosis detection, which was in line with our own findings \cite{2759-05}. Due to this, it seems interesting to have a first stage identifying mitotic figures, which acts as a coarse classifier and do the fine classification in a secondary stage, where domain adaption is most crucial. Thus, this work focuses on the distinguishing of mitotic figures from similar looking cells (hard negatives), which would be the task of the second stage.

\section{Material and Methods}
\begin{figure}[b]
	\centering
	\includegraphics[width=\textwidth]{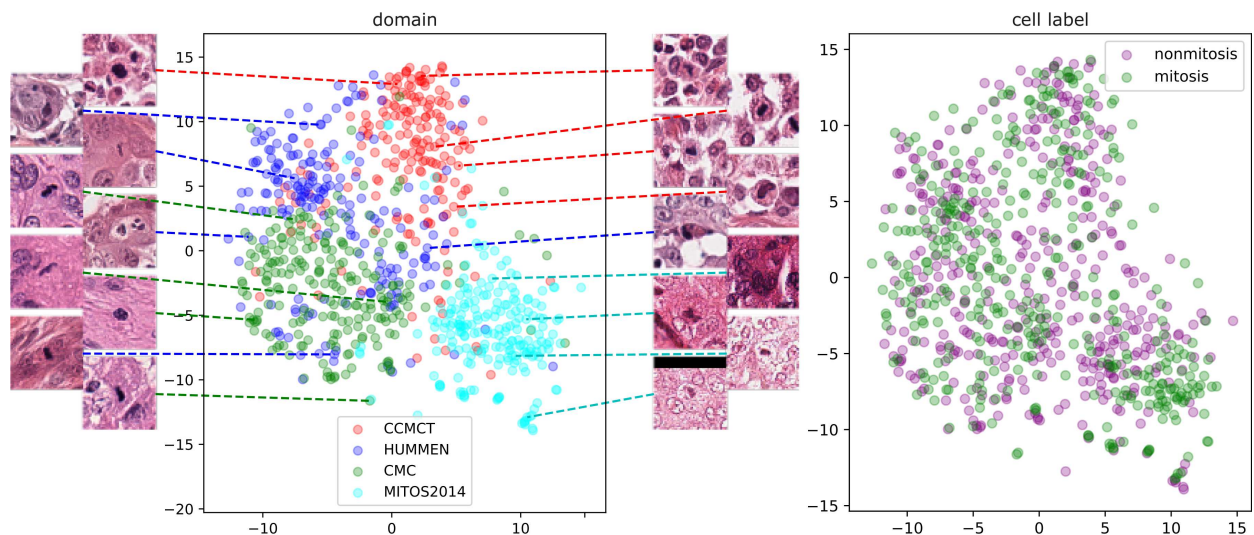}
	\caption{T-SNE representation of the various domains and classes in our data set, features extracted from a ResNet-18 stem, trained on ImageNet. Human data sets are from mammary carcinoma (MITOS2014) and meningioma (HUMMEN), canine data sets are from cutaneous mast cell tumor (CCMCT) and mammary carcinoma (CMC).}
\end{figure}

In this work, we use four data sets, two from canine histopathology slides and two from human histopathology slides. Our research group recently published a data set of canine cutaneous mast cell tumor (CCMCT, a common tumor in dogs), spanning 32 whole slide images and providing annotation for all mitotic figures on the complete slides \cite{2759-05}. 
For this work, we have generated additionally a data set of canine mammary carcinoma (CMC) which includes 22 tumors also completely annotated blindly by two expert pathologists and with consensus found for all disagreed cells using an open source solution\cite{2759-11}. 
As for human tissue, we set up a pilot data set of human meningioma (HUMMEN) consisting of three whole slide images from WHO grades I, II and III.  All tissue was sampled for routine diagnostics and, where applicable, institutional review board approval and written consent was given (IRB approval number hidden for review). Additionally, to set a baseline, we used the publicly available MITOS2014 data set \cite{2759-03}, which contains mitotic figure annotations on human mammary tissue.
All four data sets have in common that they contain not only true mitotic figure annotations, but also annotations that can easily be misinterpreted by either an algorithm or a human, but were in final consensus of all experts not found to be mitotic figures. These can be considered truly hard negative, and thus the differentiation between this group and the group of mitotic figures is one of the hardest tasks in mitotic figure detection.  As Fig. \ref{datasets} shows, the domain shift between the data sets can be nicely be spotted in a t-SNE representation, while the cell label is not easily distinguishable. 

\begin{table}[t]
\caption{Comparison of the data sets used in this study.}
\resizebox{\textwidth}{!}{
\begin{tabular}{|l|l|l|r|r|r|r|}
\hline
    data set name & species & tumor & cases &  mitotic figures & mitotic figure look-alikes & annotations available\\
    \hline
	CCMCT \cite{2759-05} & dog & cutaneous mast cell tumor & 32 & 44,880 & 27,965 & complete WSI \\
	CMC (unpublished) & dog & mammary carcinoma & 22 & 13,385 & 26,585 & complete WSI \\
	MITOS 2014 \cite{2759-03} & human &  mammary carcinoma & 11 & 749 & 2,884 & selected area\\
	HUMMEN (unpublished) & human & meningioma & 3 & 782 & 1,721 & complete WSI\\
	\hline
\end{tabular}}	
\label{datasets}
\end{table}


Domain-adversarial training \cite{2759-06} is an unsupervised network adaptation method, recently also employed in the field of histopathology \cite{2759-07}. It aims to account for a domain shift in latent space that is known to occur between slides of the same data set, but also, and more severely, between data sets. 
The core idea is to train a model with a body and two heads, where one head aims to perform classification of the cell class, and the other would aim for the classification of the domain. The latter inherently uses said domain shift to differentiate between data sets, however, it is this domain shift that we want to reduce. 

\begin{figure*}[t]
	\caption{Domain adversarial training of our network.}
	\includegraphics[width=\textwidth]{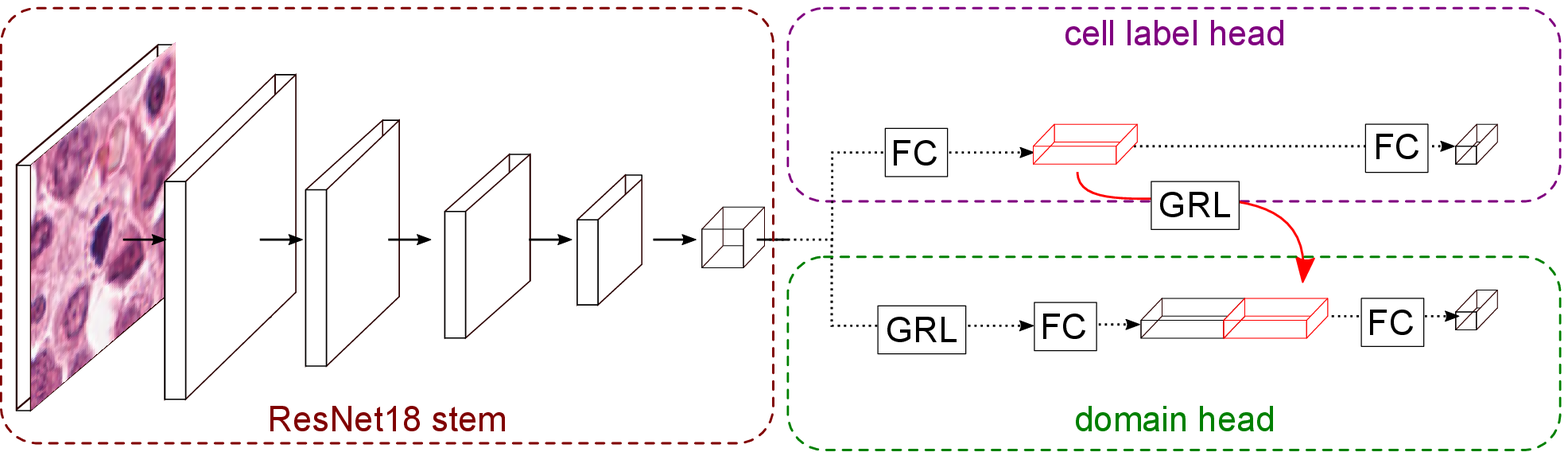}
	\label{fig:approach}
\end{figure*}

Ganin \etal proposed to use gradient reversal during network training for this task \cite{2759-06}. The gradient reversal layer introduced by them acts as a pure forwarding layer during inference, but inverts the sign of all gradients during back-propagation. 
 We aim to use the method to make the latent space representations for cell classification indistinguishable between source and target domain, and thus improve classification for the target domain in an unsupervised way. Using squared input images of 128px size with the respective cell at it's center, we employed a state-of-the-art network stem (ResNet-18 \cite{2759-08}) for the main feature extraction. After flattening the feature vector, we have two linear (fully connected) layers each with batch-norm and ReLU activation after the first layer as the cell label head (Fig. \ref{fig:approach}). The secondary domain classification head starts with a gradient reversal layer (GRL) and another fully connected layer with batch normalization. Afterwards, we concatenate the intermediate features of the classification head into the secondary branch, to also be able to reduce domain shift in this layer (as proposed by Kamnitsas \etal \cite{2759-09}).

\subsection*{Training}
For training, we feed all images, and suppress the cell label information (mitosis/nonmitosis) in the loss function for samples of the target domain.
We did not use a validation set for model selection, as a train/val/test split on patient level for that purpose is questionable as the domain shift will likely be also high within the data set. 
We chose an even distribution of cell labels and domains for both, the training and the test set.  

We denote the output probability of the cell classification branch and the domain classification branch $p_C$ and $p_D$, respectively. The true cell class and domain are denoted as $y_C$ and $y_D$, respectively. We derive the domain classification loss as simple cross-entropy loss:

\begin{equation}
	L_\textrm{D}(y_D, p_D) = \left\{ \begin{array}{lll} -log(p_D)& \hspace{1cm} & \textrm{if~} y_D = 1 \\
 - log(1-p_D) & & \textrm{otherwise}	\end{array}
\right.
\end{equation} 

With $y_D=1$ representing the target domain, the cell classification loss is:
\begin{equation}
	L_\textrm{C}(y_C, p_C, y_D) = \left\{ \begin{array}{lll} -log(p_D)& \hspace{1cm} & \textrm{if~} y_C = 1 \textrm{~and~} y_D = 0\\
 - log(1-p_D) & & \textrm{if~} y_C = 0 \textrm{~and~} y_D = 0	\\
 0 & & \textrm{otherwise}\end{array}
\right.
\end{equation} 

The total loss is a normalized weighted sum of both, i.e with uppercase symbols denoting the respective mini batch vectors
\begin{equation}
 L = \frac{\gamma}{\sum_i (1 - Y_{D,i})} \sum_{i} L_\textrm{C}(Y_{C,i}, P_{C,i}, Y_{D,i}) + \sum_{i} L_\textrm{D} (y_D, p_D)
\end{equation}
where $\gamma$ is a tunable parameter. We trained for 3x30 epochs using super-con\-vergence~\cite{2759-10} and Adam as optimizer. For better comparability, we limited the number of cells for the large data sets to 1,600.

\section{Results}

\begin{figure}[hbt]
\centering
	\subfigure[Box-whisker plots of cell label accuracy]{\includegraphics[height=8cm]{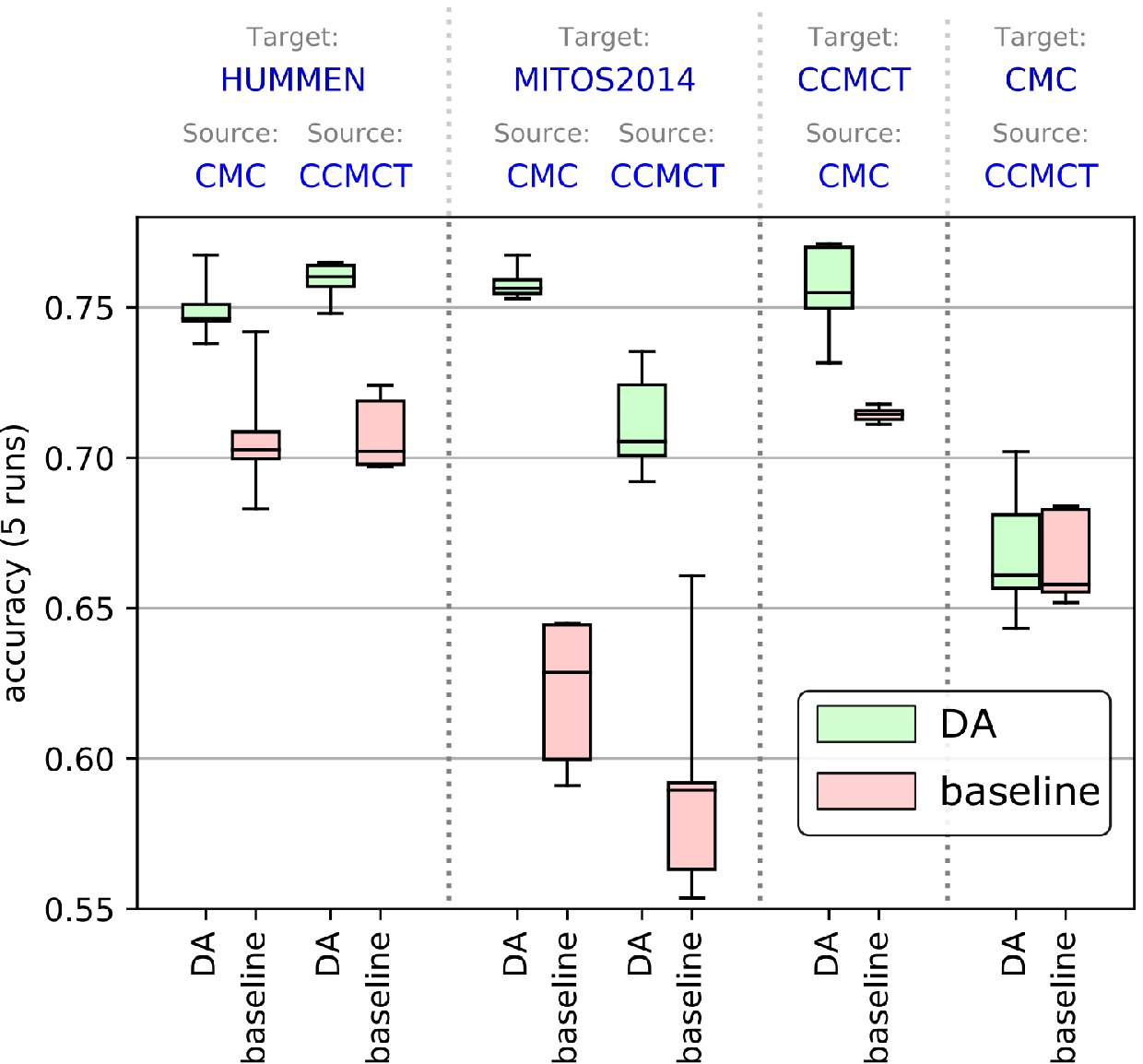}
	    \label{fig:results}}
	    \hspace{0.5cm}
	\subfigure[Feature vector (t-SNE)]{\includegraphics[height=8cm]{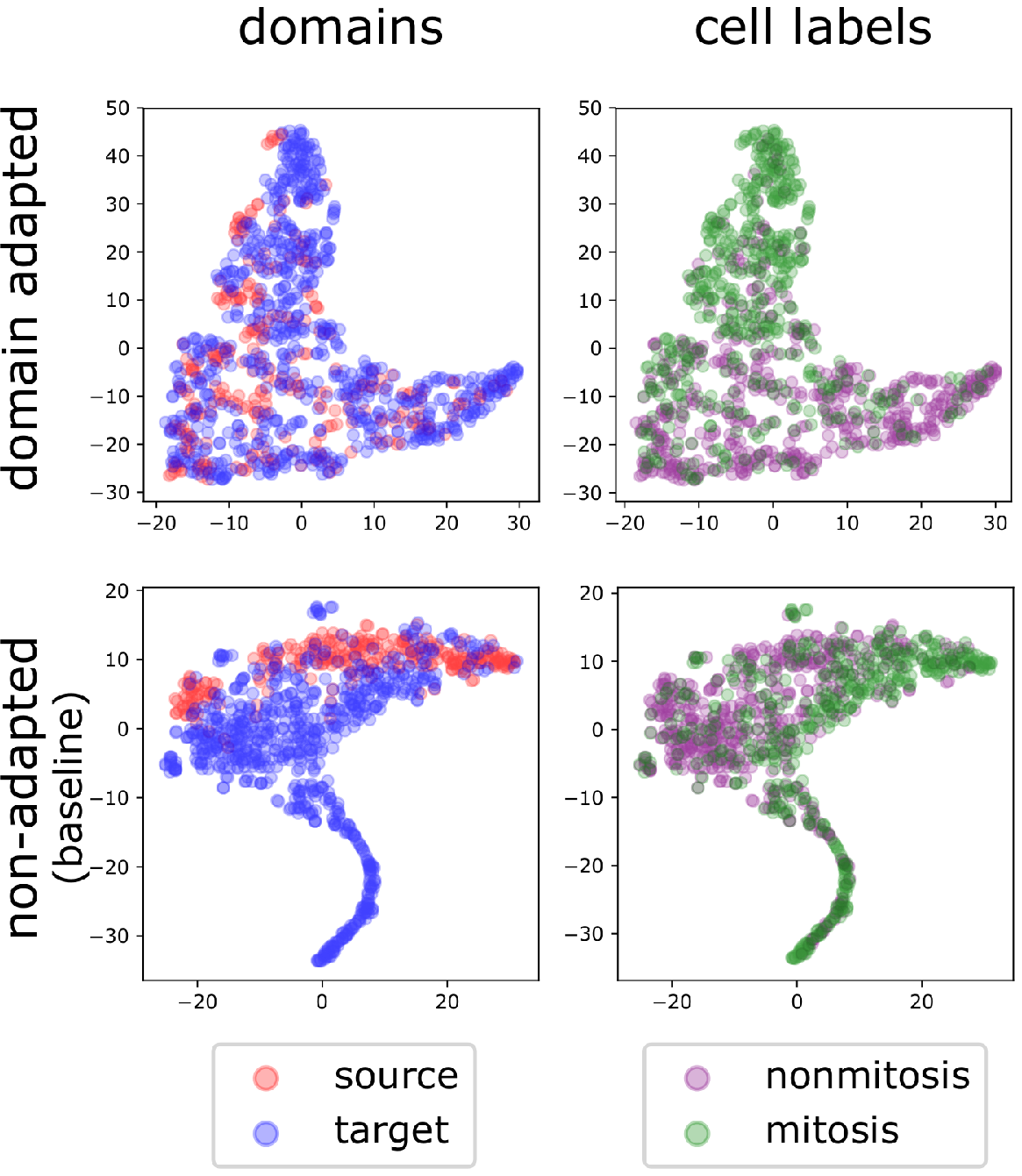} 
	     \label{fig:tsne}}
	\caption{Results of domain transfer. Plot a) shows accuracy for 5 runs on the different domain transfer tasks, whereas b) shows an example ResNet-18-feature vector with and without domain transfer (CCMCT to MITOS2014 task). }
\end{figure}
Domain transfer raised the mean accuracy over 5 runs to on the HUMMEN data set by 4.7 percentage points, on the MITOS2014 data set by 12.8 percentage points and in total by 6.5 percentage points (Fig. \ref{fig:results}).
This improvement can largely be attributed to a reduced domain shift in latent space (Fig. \ref{fig:tsne}).
The performance varied both for the domain-adapted and the non-adapted case, and was especially large, where the original domain shift was most severe (MITOS2014 target data set). In this case, we achieved the best results when training on the same tissue type for a different species (CMC).

\section{Discussion}
Our results indicate that domain adversarial training is a suitable unsupervised learning method to perform mitotic figure domain adaptation between tissue and, equally importantly, also species. This is especially beneficial as publicly available, large datasets can be used for new domains - regardless of tissue/tumor type or species (animal or human). Further, we find a clear dependency on source and target data set, which is not surprising considering the initial latent space distributions. 

While the results showed clear benefits for the given task, we should point out our method needs access to pre-selected mitotic figures candidates, which we extracted beforehand. This step could, however, be taken over by generalizing model that was trained on a broad range of tissue, which we aim to investigate in future research.

\subsection*{Acknowledgements}
CAB gratefully acknowledges financial support received from the Dres. Jutta \& Georg Bruns-Stiftung f\"ur innovative Veterin\"armedizin.
\bibliographystyle{unsrt}

\end{document}